\documentclass[10pt,final,journal,twocolumn]{IEEEtran}

\usepackage{booktabs}
\usepackage{amsmath}
\usepackage{amsthm}
\usepackage{mathrsfs}
\usepackage{mathtools}
\usepackage{subfigure}
\usepackage{hyperref}
\usepackage{graphicx}
\usepackage{latexsym}
\usepackage{amssymb,amsbsy,verbatim,array}
\usepackage{epsfig}
\usepackage{epstopdf}
\usepackage{cite}
\usepackage{color}
\usepackage{algpseudocode}
\usepackage{algorithmicx, algorithm}
\usepackage{makecell}
\usepackage{extarrows}
\usepackage{multicol}
\usepackage{bm}
\usepackage{mathrsfs}

\begin{document}

\title{Towards Optimally Efficient Search with Deep Learning for Large-Scale MIMO Systems}

\author{Le He, Ke He, Lisheng Fan, Xianfu Lei, Arumugam Nallanathan, \emph{Fellow, IEEE} \\ and George K. Karagiannidis, \emph{Fellow, IEEE}
    \thanks{L. He and L. Fan are both with the School of Computer Science and Cyber Engineering, Guangzhou University, China (e-mail: hele20141841@163.com, lsfan@gzhu.edu.cn).}
    \thanks{K. He was with the School of Computer Science and Cyber Engineering, Guangzhou University, China, and is  now with the Signal Processing \& Satellite Communications Research Group (SIGCOM), Interdisciplinary Centre for Security, Reliability and Trust (SnT) - University of Luxembourg, L-1855 Luxembourg. (e-mail: heke2018@e.gzhu.edu.cn, ke.he@uni.lu).}
    \thanks{X. Lei is with the School of Information Science and Technology, Institute
        of Mobile Communications, Southwest Jiaotong University, Chengdu 610031,
        China (e-mail: xflei@home.swjtu.edu.cn).}
    \thanks{A. Nallanathan is with the School of Electronic Engineering and Computer Science, Queen Mary University of London, London, U.K. (e-mail: a.nallanathan@qmul.ac.uk)}
    \thanks{G. K. Karagiannidis is with the Wireless Communications and Information Processing Group (WCIP), Aristotle University of Thessaloniki, Thessaloniki 54 124, Greece (e-mail: geokarag@auth.gr).}}

\pagestyle{headings}

\maketitle

\thispagestyle{empty}

\begin{abstract}
    This paper investigates the optimal signal detection problem with a particular interest in large-scale multiple-input multiple-output (MIMO) systems. The problem is NP-hard and can be solved optimally by searching the shortest path on the decision tree. Unfortunately, the existing optimal search algorithms often involve prohibitively high complexities, which indicates that they are infeasible in large-scale MIMO systems. To address this issue, we propose a general heuristic search algorithm, namely, hyper-accelerated tree search (HATS) algorithm. The proposed algorithm employs a deep neural network (DNN) to estimate the optimal heuristic, and then use the estimated heuristic to speed up the underlying memory-bounded search algorithm. This idea is inspired by the fact that the underlying heuristic search algorithm reaches the optimal efficiency with the optimal heuristic function. Simulation results show that the proposed algorithm reaches almost the optimal bit error rate (BER) performance in large-scale systems, while the memory size can be bounded. In the meanwhile, it visits nearly the fewest tree nodes. This indicates that the proposed algorithm reaches almost the optimal efficiency in practical scenarios, and thereby it is applicable for large-scale systems. Besides, the code for this paper is available at \url{https://github.com/skypitcher/hats}.
\end{abstract}

\begin{IEEEkeywords}
    Signal detection, integer least-squares, deep learning, maximum likelihood detection, MIMO, sphere decoding, best-first search
\end{IEEEkeywords}

\section{Introduction}
Given an observation vector $\bm{y}\in\mathbb{R}^{n \times 1}$ and the transformation matrix $\bm{H} \in \mathbb{R}^{n \times m}$ with full column rank ($n \geq m$), the signal detection problem aims to recover the transmitted discrete signal $\bm{x} \in \mathbb{R}^{m \times 1}$ from the linear mixing model
\begin{align}\label{Eq::SysReal}
    \bm{y} = \bm{H}\bm{x} + \bm{w},
\end{align}
where $\bm{w} \in \mathbb{R}^{n \times 1}$ denotes the additive white Gaussian noise with zero mean and unit variance. Generally, the detection problem has diverse applications which include -but it is not limited to- signal processing  \cite{HassibiV05}, communications  \cite{SommerFS05}, machine learning  \cite{Bishop07}, global navigation satellite systems (GNSS)  \cite{ teunissen2010integer}, and radar imaging  \cite{goldberger2009gaussian}. In these areas, estimating the transmitted signal optimally in large-scale systems still remains a big challenge, and it is also of vital importance for improving the reliability of the whole system  \cite{YangH15}. In this paper, we will henceforth concentrate on studying the optimal signal detection in large-scale multiple-input multiple-output (MIMO) wireless communication systems. However, it shall be noted that the proposed method is general, and can be applied to other applications.

A large-scale MIMO system  \cite{SanguinettiBH20,BjornsonSWHM19,LuLSAZ14}, which is one of the most promising wireless techniques to significantly improve the spectrum efficiency, has been widely studied during the last decades. Specifically, there are $m_c$ and $n_c$ antennas equipped at the transmitter and receiver such that $n_c \geq m_c \gg 1$. Moreover, $\bm{H}_c$ refers to the random wireless channel state information (CSI) matrix with the $(i,j)$'s element $h_{i,j} \sim \mathcal{CN}(0, \rho)$ denoting the tap gain from the $j$-th transmit antenna to the $i$-th receive antenna. In further, $\bm{x}_c$ denotes the transmit signal vector which is uniformly distributed over the finite set of lattice points $\mathcal{A}^m$. In fact, $\mathcal{A}$ is the alphaset depending on the underlying modulation schema, and a standard example of $\mathcal{A}$ would be $\mathcal{A}=\{a+bj| a,b \in \{-1, +1\}\}$ for 4-quadrature amplitude modulation ($4$-QAM). While transmitting $\bm{x}_c$ through the random wireless channel $\bm{H}_c$, it suffers from an additive white Gaussian noise $\bm{w}_c$. Therefore, the received signal can be expressed as
\begin{align}\label{Eq::SysComplex}
    \bm{y}_c = \bm{H}_c\bm{x}_c + \bm{w}_c.
\end{align}
In particular, we consider the following equivalent representation
\begin{align}
     & \bm{y} = \begin{bmatrix}
        \Re(\bm{y}_c) \\
        \Im(\bm{y}_c)
    \end{bmatrix},
     &                                         & \bm{H} = \begin{bmatrix}
        \Re(\bm{H}_c) & -\Im(\bm{H}_c) \\
        \Im(\bm{H}_c) & \Re(\bm{H}_c)  \\
    \end{bmatrix}, \\
     & \bm{x} = \begin{bmatrix}
        \Re(\bm{x}_c) \\
        \Im(\bm{x}_c)
    \end{bmatrix},
     &                                         & \bm{w} = \begin{bmatrix}
        \Re(\bm{w}_c) \\
        \Im(\bm{w}_c)
    \end{bmatrix},
\end{align}
where $\bm{y}_c \in \mathbb{C}^{n_c \times 1}$, $\bm{H}_c \in \mathbb{C}^{n_c \times m_c}$, $\bm{x}_c \in \mathbb{C}^{m_c \times 1}$ and $\bm{w}_c \in \mathbb{C}^{n_c \times 1}$ are composed of complex values, and $\Re(\cdot)$ and $\Im(\cdot)$ denote the real part and imaginary part of the value, respectively. Based on this real-valued representation, (\ref{Eq::SysComplex}) can be written as (\ref{Eq::SysReal}) with $m = 2m_c$ and $n = 2n_c$.

In lattice theory, $\bm{H}$ is considered to be the generator matrix  \cite{AgrellEVZ02} of the generated lattice $\mathcal{L}(\bm{H}) = \{ \bm{H}\bm{x} | \bm{x} \in \mathcal{A}^m \}$, which indicates that the generated lattice is the ``skewed'' one of the original lattice. When $\bm{H}$ is perfectly known at the receiver, the mathematically optimal approach to solve the detection problem in terms of minimizing the average error probability, is to search the closest ``skewed'' lattice point $\bm{H}\bm{x}^*$ to $\bm{y}$, in terms of the Euclidean distance\cite{SommerFS05}
\begin{align}\label{Eq::MLD}
    \bm{x}^* = \mathop{\arg\min}_{\hat{\bm{x}} \in \mathcal{A}^m} \left\|\bm{y}-\bm{H}\hat{\bm{x}}\right\|^2,
\end{align}
which gives the exact optimal maximum likelihood (ML) estimate of $\bm{x}$ in (\ref{Eq::SysReal}). Since the signal vector $\bm{x}$ comprises of integer components only, (\ref{Eq::MLD}) is also refereed as integer least-squares (ILS) problem in the literature  \cite{ZhaoG06}. Unfortunately, solving the ILS problem is much more challenging with comparison to the standard least-squares problem, where the latter's signal vector comprises of continuous entries rather than discrete entries, and the optimal solution can be efficiently resolved via pseudo inverse  \cite{HassibiV05}. As a combinatorial optimization problem, it is known to be NP-hard because of its discrete search space. It has been proven that solving the problem optimally involves exponential complexity for all algorithms in the worst case  \cite{Micciancio01,ChenPWLTXW15}. Moreover, with the deployment of 5G, the numbers of antennas $m_c$ and $n_c$ become very large to support the demand of high data rate and ultra-reliable low-latency communication (URLLC), which indicates that finding the optimal estimate becomes much more challenging as well.  However, one can formulate the problem as a search in a state space, and efficiently reduce the average complexity by following a certain strategy that investigates only the necessary lattice points at each dimension \cite{HassibiV05}. In particular, the resulting state space forms a decision tree whose nodes represent decisions on symbols, and branches represent the costs of the associated decisions. Correspondingly, the strategies that decide which node should be chosen for expansion are called tree search algorithms. Once we find the shortest path, we can find the optimal solution.

\subsection{Related Works}
In the past decades, researchers have proposed many tree search algorithms to address the ILS problem  \cite{AgrellEVZ02, SommerFS05, ZhaoG06, LiuLHYG19, DBLP:journals/jsac/GuoN06, LiuLMP21, BaroHW03,YangLH05, MuruganGDC06, CuiHT06, DaiY09, ChangCL12}, and a comprehensive survey of these algorithms can be found in the fifty-years review of MIMO detection  \cite{YangH15}. One common approach achieving the optimal ML detection performance is the sphere decoding (SD) algorithm, which employs a branch-and-bound (BnB) depth-first search (DFS) strategy to find the shortest path  \cite{AgrellEVZ02, SommerFS05, ZhaoG06}. Besides, many variants of SD have been proposed in the literature as well. Among these variants, the Schnorr-Eucherr SD (SE-SD) achieves the optimal performance with reduced average complexity \cite{AgrellEVZ02}. However, its average complexity still remains prohibitively high in large-scale MIMO systems with low signal-to-noise ratio (SNR) \cite{HassibiV05}. To overcome this drawback, $K$-best SD has been proposed to achieve a fixed and reduced complexity with a breath-first search (BrFS) strategy, whereas the performance is sacrificed  \cite{DBLP:journals/jsac/GuoN06}. More importantly, the complexity reduction can not be guaranteed in high SNR regimes  \cite{9425021}.

Attempts that employ best-first search (BeFS) or stack algorithms to overcome the drawbacks of DFS and BrFS have been investigated in the literature  \cite{Russell92, HanHC93, BaroHW03,YangLH05, MuruganGDC06, CuiHT06, DaiY09, ChangCL12}. Among these variants, A* like BeFS is the most promising variant which employs a heuristic function to predict the shortest remaining cost  \cite{BaroHW03}. It guarantees to find the optimal solution with an admissible heuristic function, while it achieves the optimal efficiency with a consistent heuristic function \cite{DechterP85}. It has been show that BeFS visits the fewest nodes among the three search strategies, but it requires exponentially increasing memory space  \cite{MuruganGDC06}. Fortunately, several memory-bounded BeFS algorithms have been proposed in the past years, and their performances are very close to the original one  \cite{Russell92, HanHC93, DaiY09}. However, it is still very difficult to find a consistent heuristic function such that the optimal efficiency is achieved  \cite{CuiHT06, ChangCL12}. Hence, the existing BeFS algorithms are still infeasible for large-scale MIMO systems.

With the tremendous success of machine learning techniques on physical layer communication, significant improvement becomes possible for the aforementioned search strategies  \cite{Kim20, XiaHXZFK20, DBLP:journals/tcom/HeHFDKN21, HeWJL20, XueBCDYRS18,LinGZZA21}. For example, the authors in   \cite{MaZLGJ19} proposed a novel expectation maximization-based sparse Bayesian learning framework  to learn the model parameters of the sparse virtual channel, which significantly reduces the overhead of the channel training in massive MIMO systems.  To solve the crucial active antenna selection problem in massive MIMO, a deep learning (DL) based active antenna selection network was devised to utilize the probabilistic sampling theory to select the optimal location of these active antennas  \cite{ZhangZGMD21}. Besides, a preliminary theoretical analysis on DL based channel estimation was presented in  \cite{HuGZJL21}  to understand and interpret the internal mechanism of single-input multiple-output (SIMO) systems. The theoretical result shows that DL based channel estimation outperforms or is at least comparable with traditional channel estimation, depending on the types of channels, which encourages researchers to solve problems with promising DL based approaches. Regarding the ILS problem, researchers have proposed a DL based SD (DL-SD) algorithm which significantly improves SE-SD's performance by employing the deep neural networks (DNNs) to choose a good initial radius  \cite{AskriO19, MohammadkarimiM19}. Since the tree search can be treated as a sequential decision making process, the authors in  \cite{Sun20} proposed a sub-optimal search algorithm named LISA with a fixed complexity, since LISA uses a DNN to make decisions with fixed steps. It was demonstrated that LISA could achieve very good performances with a fixed complexity. The success of these data-driven search algorithms certainly motivates us to improve the BeFS algorithm's performance with DL techniques.

\subsection{Contributions}
Consequently, we are interested in speeding up A* like BeFS algorithms with model-driven DL methods, while trying not to compromise the optimal bit error rate (BER) performance. To accomplish this goal, we propose a hyper-accelerated tree search (HATS) algorithm, which enhances the underlying efficient memory-bounded A* algorithm (SMA*)  \cite{Russell92}, through predicting the optimal heuristic with a well-trained DNN. This idea is inspired by the fact that with the optimal heuristic, A* not only becomes optimally efficient but also expands the fewest nodes  \cite{DechterP85}. In contrast, the other A* inspired search algorithms only find the admissible heuristic  \cite{CuiHT06, ChangCL12}, which limits the improvement of these methods. As we will show in the simulations, the proposed algorithm achieves almost the optimal BER performance in large-scale MIMO systems, while its memory size can be bounded and it reaches the lowest average complexity under practical scenarios. This suggests that the proposed algorithm is feasible for practical large-scale MIMO systems.

\section{MIMO Signal Detection with Heuristic Best-First Search}
In this section, we will take an efficient approach to interpret the optimal MIMO signal detection problem as a tree search, which enables us to decouple the vector-valued problem into a sequential decision making problem. After that, we will introduce the A* algorithm to find the shortest path, and discuss its properties accordingly.

\subsection{Tree Construction}
\begin{figure}[t!]
    \centering
    \includegraphics[width=3.25in]{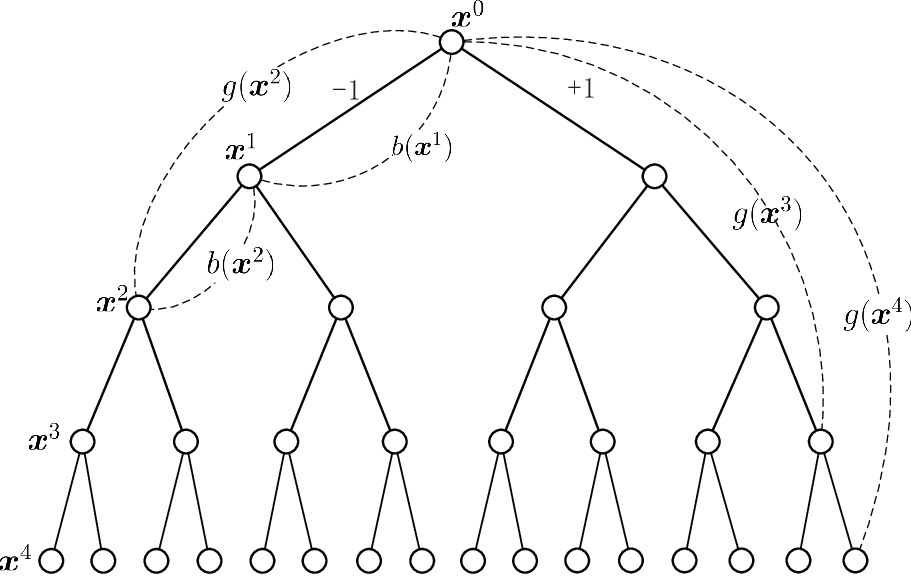}
    \caption{An example of decision tree for signal detection, where $m=4$ and $\mathcal{A}=\{-1, +1\}$. In the resulting tree, the left branch represents the symbol decision of  $-1$, while the right branch represents the symbol decision of $+1$.}
    \label{Fig::Tree}
\end{figure}

To construct a decision tree, we first perform QR decomposition on the CSI matrix $\bm{H}$ as
\begin{align}\label{eq:QRD}
    \bm{H} = \begin{bmatrix}\bm{Q}_1 & \bm{Q}_2 \end{bmatrix}
    \begin{bmatrix}
        \bm{R} \\
        \bm{0}_{(n-m) \times m}
    \end{bmatrix}
    = \bm{Q}_1\bm{R},
\end{align}
where $\bm{R} \in \mathbb{R}^{m \times m}$ is an upper triangular matrix, and the partitioned matrices $\bm{Q}_1 \in \mathbb{R}^{n \times m}$  and $\bm{Q}_2 \in \mathbb{R}^{n \times (n-m)}$  both have orthogonal columns. Then, we rewrite (\ref{Eq::SysReal}) as
\begin{align}\label{Eq::SysQR}
    \bm{z} = \bm{R}\bm{x} + \bm{v},
\end{align}
where $\bm{z} \triangleq \bm{Q}_1^T\bm{y}$ and $\bm{v} \triangleq \bm{Q}_1^T\bm{w}$. Note that the optimal MIMO signal detection problem is always squared of dimensions $m$ after preprocessing. For convenience, we number the entries of matrices and vectors of (\ref{Eq::SysQR}) in a reverse order as
\begin{align}
    \begin{bmatrix}
        z_m    \\
        \vdots \\
        z_1
    \end{bmatrix}
    =
    \begin{bmatrix}
        r_{m,m} & \hdots & r_{m, 1} \\
        \vdots  & \ddots & \vdots   \\
        0       & \hdots & r_{1,1}
    \end{bmatrix}
    \begin{bmatrix}
        x_m    \\
        \vdots \\
        x_1
    \end{bmatrix}
    +
    \begin{bmatrix}
        v_m    \\
        \vdots \\
        v_1
    \end{bmatrix},
\end{align}
where $r_{i,j}$ represents the ($i, j$)-th component of $\bm{R}$ after arranging from the bottom right to the upper left. After preprocessing, the squared Euclidean distance of a given candidate can be expanded as
\begin{subequations}
    \begin{align}
        d^2(\bm{x}) = & \underbrace{(z_1 - r_{1,1}x_1)^2}_{b(\bm{x}^1)} +\underbrace{(z_2 - r_{2,2}x_2 - r_{2,1}x_1)^2}_{b(\bm{x}^2)}+ \cdots  \nonumber \\
                          & + \underbrace{(z_{m} - r_{m,1}x_{1} -  \cdots - r_{m,m}x_m)^2}_{b(\bm{x}^m)}                                                         \\
        =                 & \sum_{k=1}^m{b(\bm{x}^k)},
    \end{align}
\end{subequations}
where the $k$-th incremental cost is denoted by
\begin{align}
    b(\bm{x}^k) = \left(z_{k} - \sum_{j=1}^{k} r_{k,j}x_{j}\right)^2,
\end{align}
which only depends  on the partial signal vector (PSV) $\bm{x}^k = [x_k, x_{k-1}, \hdots, x_1]^T$. The cumulative cost of the PSV $\bm{x}^k$ is
\begin{align}
    g(\bm{x}^k) = \sum_{i=1}^{k} b(\bm{x}^i),
\end{align}
and the successor can be computed recursively by
\begin{align}\label{Eq::TreeRecursive}
    g(\bm{x}^{k+1}) = b(\bm{x}^{k+1}) + g(\bm{x}^{k}), \quad \forall k = 0, 1, \cdots, m-1
\end{align}
where the initialization $g(\bm{x}^{0}) \equiv 0$ is set. Clearly, (\ref{Eq::TreeRecursive}) constructs a perfect $\vert\mathcal{A}\vert$-way $m$-level (starts from zero) decision tree. In the resulting tree, the deepest nodes represents the goal nodes associated with specific candidates. Since $\vert\mathcal{A}\vert$ is finite, the set of goal nodes can be enumerated as
\begin{align}
    \mathcal{A}^m = \{\bm{x}^m_{1}, \bm{x}^m_{2}, \hdots, \bm{x}^m_{j}, \hdots, \bm{x}^m_{\vert\mathcal{A}\vert^m}\},
\end{align}
where each goal node $\bm{x}^m_{j}$ is associated with a signal candidate. For convenience, we denote the antecedent of a goal node $\bm{x}^m_j$ that locates at level $k$ as $\bm{x}^k_j$, which also represents the path leading from the root to that node. Note that we will sometimes ignore the subscript and use $\bm{x}^k$ to represent an arbitrary node located at level $k$. We also use $\bm{x}^j \subseteq \bm{x}^k$ ($0 \leq j \leq k$) to denote that $\bm{x}^l$ is the antecedent of $\bm{x}^k$. On the contrary, $\bm{x}^j \supseteq \bm{x}^k$ ($0 \leq k \leq j$) denotes that $\bm{x}^j$ is the descendent of $\bm{x}^k$. Importantly, each node $\bm{x}^k$ is associated with a branch cost $b(\bm{x}^k)$, and $g(\bm{x}^k)$ is the cumulative cost of the path leading from the root to that node. For demonstration, a simple binary $4$-level decision tree is illustrated in Fig. \ref{Fig::Tree}, where $m=4$ and $\mathcal{A}=\{-1, +1\}$.

\subsection{Heuristic Tree Search}
Now it is straightforward to employ the A* algorithm  \cite{DechterP85,CuiHT06, ChangCL12, HeKeTCOMGSM} to search the least-cost path on the tree. Before describing the algorithm, it is necessary to introduce the following notations,
\begin{itemize}
    \item \textsf{ACTIVE} is an \emph{ordered list with possibly limited space} that stores nodes to be expanded, whose data structure is often a priority queue or self-balancing binary search tree (a.k.a. AVLTree) in practice. In this case, we can efficiently find the least-$f$-cost node with a computational complexity of $\mathcal{O}(1)$, which means that the algorithm is able to find the least-f-cost node at each loop with a very small fixed cost.
    \item $f(\bm{x}^k)$ is the \emph{evaluation cost ($f$-cost)} currently assigned to a node $\bm{x}^k$ at the time. Note that $f(\bm{x}^k)$ is not static, and it may change during search.
    \item A node is \emph{visited} or \emph{generated} if it has been encountered during the expansion of its parent.
    \item A node is \emph{expanded} if all of its successors have been generated during the expansion.
    \item A node is \emph{in memory} if it is inserted into \textsf{ACTIVE}.
    \item The \emph{complexity} of a search algorithm is defined in terms of the number of visited nodes.
\end{itemize}
We are now ready to introduce the A* algorithm, and the pseudo code is presented in Algorithm  \ref{Algo::AStar}. Throughout the paper, we shall emphasize that the tree search always starts at the dummy root $\bm{x}^0$. At each iteration, A* expands the least-$f$-cost node among all the nodes that are in $\textsf{ACTIVE}$ and waiting for expansion, and inserts all successors of that node into $\textsf{ACTIVE}$ afterwards  \cite{ChangCL12}. This process will terminate as long as a goal node is selected for expansion, and the selected goal node will become the output of the algorithm. As a heuristic BeFS algorithm, A* employs a heuristic function $h(\bm{x}^k)$ to guide the search. Specifically, the heuristic function estimates the remaining cost from $\bm{x}^k$ to goal nodes, denoted by
\begin{align}\label{Eq::RemainCost}
    h(\bm{x}^k, \bm{x}^m) &= g(\bm{x}^m) - g(\bm{x}^k) \\
    &= \sum_{i=k+1}^{m}\left(z_{i} - \sum_{j=1}^{i} r_{i,j}x_{j}\right)^2, \quad \forall \bm{x}^m \supseteq \bm{x}^k.
\end{align}
Accordingly, A* computes the evaluation cost of $\bm{x}^k$ as
\begin{align}
    f(\bm{x}^k)=g(\bm{x}^k)+h(\bm{x}^k),
\end{align}
where $h(\bm{x}^k)$ represents the estimate of the minimum value of (\ref{Eq::RemainCost}), and $h(\bm{x}^m) = 0$ holds since the goal nodes have no successor at all. Therefore, $f(\bm{x}^k)$ actually estimates the cost of the shortest path of the sub-tree of $\bm{x}^k$. In particular, the optimal heuristic function always gives the minimum value of (\ref{Eq::RemainCost}) as
\begin{align}\label{Eq::OptimalHeuristic}
    h^*(\bm{x}^k) = \min_{\bm{x}^m \supseteq \bm{x}^k} h(\bm{x}^k, \bm{x}^m).
\end{align}
In this case, the optimal $f$-cost is given by
\begin{align}\label{Eq::OptimalCost}
    f^*(\bm{x}^k) = g(\bm{x}^k) + h^*(\bm{x}^k).
\end{align}
With the optimal heuristic function, A* only expands the shortest path, as it always guides the search process towards the shortest path.

\begin{algorithm}[t!]
    \caption{A* Algorithm}\label{Algo::AStar}
    \hspace*{0.02in} { \bf Input:}{$\bm{z}$ and $\bm{R}$} \\
    \hspace*{0.02in} { \bf Output:}{$\hat{\bm{x}}$ (estimate of the transmitted signal $\bm{x}$)}
    \begin{algorithmic}[1]\State insert the root node $\bm{x}^0$ in \textsf{ACTIVE}\;
    \Loop
        \If{\textsf{ACTIVE} is empty}
            \State \Return $\hat{\bm{x}} = \varnothing$ with failure\;
        \EndIf
        \State $\bm{x}^k \gets $ \emph{least-$f$-cost node} in \textsf{ACTIVE}
        \If{$\bm{x}^k$ is a goal node}
            \State \Return $\hat{\bm{x}} = \bm{x}^k$ with success\;
        \EndIf
        \For{every successor $\bm{x}^{k+1} \supset \bm{x}^k$}
            \State $f^\prime(\bm{x}^{k+1}) \gets g(\bm{x}^{k+1})+h(\bm{x}^{k+1})$\;
            \State $f(\bm{x}^{k+1}) \gets \max\left\{f(\bm{x}^k), f^\prime(\bm{x}^{k+1})\right\}$\;
            \State Insert $\bm{x}^{k+1}$ into \textsf{ACTIVE}
        \EndFor
        \State Remove $\bm{x}^k$ from \textsf{ACTIVE}\;
    \EndLoop
    \end{algorithmic}
\end{algorithm}

\subsection{Optimality and Optimal Efficiency}\label{Sec::OptimalEfficiency}
We know that A* performs best with the optimal heuristic function, since only the nodes lying along the shortest path would be expanded eventually. In practical, we however often use a lower bound on (\ref{Eq::OptimalHeuristic}) as the heuristic function, as finding the optimal heuristic function requires exhaustively search as well. In general, if a heuristic function $h(\bm{x}^k)$ never overestimates the optimal heuristic, then it is \emph{admissible}  \cite{ChangCL12}. In further, $h(\bm{x}^k)$ is said to be \emph{consistent} if $h(\bm{x}^k) \leq b(\bm{x}^{k+1}) + h(\bm{x}^{k+1})$ holds for each node $\bm{x}^{k}$ and its successors $\bm{x}^{k+1}$. In other words, a heuristic function is consistent if it satisfies the triangle inequality  \cite{DechterP85}. It is clear that a consistent heuristic is also admissible, but not the vice verse.

With an admissible heuristic function, A* is said to be optimal since it guarantees to find the shortest path  \cite{DechterP85}. In particular, the simplest admissible heuristic is $h(\bm{x}^k) = 0$. Hence, the existing approaches mainly focus on finding a more meaningful admissible heuristic function based on the tighter lower bounds on (\ref{Eq::RemainCost})  \cite{CuiHT06, ChangCL12}. In this case, A* is able to find the shortest path faster than the simplest heuristic. However, the performance improvements are still limited since the estimated heuristics are still far from the optimal heuristic.

With a consistent heuristic function, A* is said to be optimally efficient in terms of that not only it is optimal but also no other optimal algorithm that uses the same heuristic would expand fewer nodes than it  \cite{DechterP85}. In this case, reducing the estimation error between a consistent heuristic and the optimal heuristic will only decrease the number of expanded nodes. Although A* can perform much better than other search algorithms, it has a main issue that its space complexity grows exponentially with the increasing search depth  \cite{DechterP85}. Consequently, finding a consistently good estimate of the optimal heuristic and improving the memory efficiency become the two essential approaches to improve the performance of A* in large-scale systems, which is also the intuition behind the strategy we will introduce in the following section.

\section{Deep Learning Accelerated Heuristic Best-First Search}
In this section, we will propose a general heuristic tree search algorithm, which significantly speeds up the heuristic BeFS algorithm by employing a DNN to estimate the optimal heuristic. After that, we will introduce the training strategy and discuss its computational complexity.

\subsection{Proposed Deep Learning Based Search Strategy}
\begin{figure}[t!]
    \centering
    \includegraphics[width=3.4in]{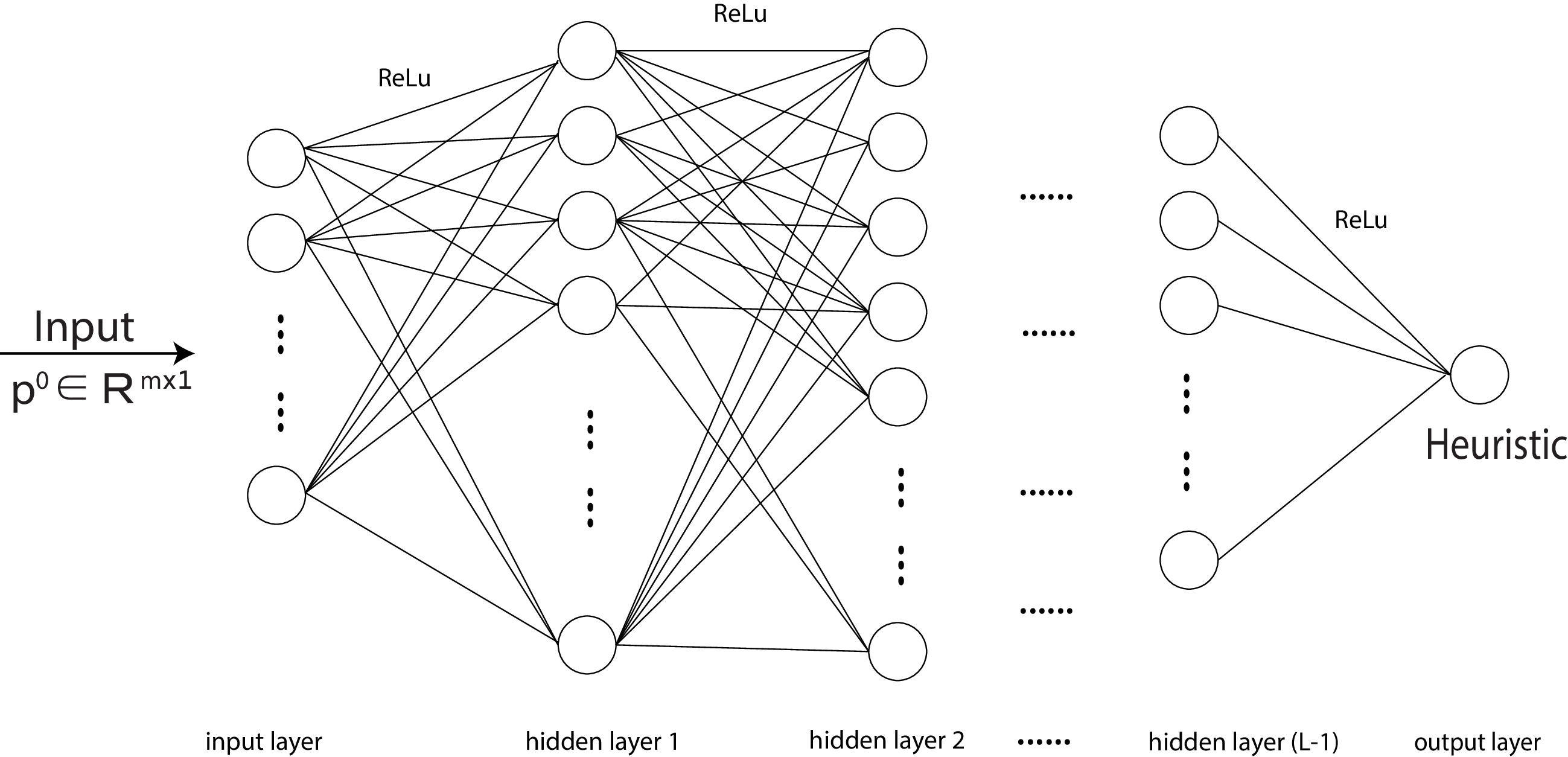}
    \caption{Network structure adopted by the proposed algorithm.}
    \label{Fig::network}
\end{figure}
Let $h(\bm{x}^k|\bm{z}, \bm{R}, \bm{\theta})$ denote the DNN parameterized by the trainable parameters $\bm{\theta}$. For convenience, we will simplify it as $h_{\bm{\theta}}(\bm{x}^k)$  in the sequel. Accordingly, we define the corresponding $f$-cost as
\begin{align}\label{Eq::DeepCost}
    f_{\bm{\theta}}(\bm{x}^k) = g(\bm{x}^k) +  h_{\bm{\theta}}(\bm{x}^k).
\end{align}
As shown in Fig. \ref{Fig::network}, we use a fully-connected neural network (FCNN) composed of $L$ fully-connected layers, and the number of neurons at the $l$-th ($1 \leq l \leq L$) layer is denoted by $n_l$. Formally, the output of the $l$-th layer will be proceeded by the rectified linear unit (ReLU) based activation function, which can be expressed as
\begin{align}
    \bm{p}^l = \max\left\{\bm{0},  \bm{W}^l\bm{p}^{l-1} + \bm{b}^l\right\},
\end{align}
where $\bm{p}^{l-1} \in \mathbb{R}^{n_{l-1}}$ is the output of the prior layer. Notations $\bm{W}^l  \in \mathbb{R}^{n_l \times n_{l-1}}$ and $\bm{b}^{l} \in \mathbb{R}^{n_l}$ are the learnable weight matrix and bias vector, respectively. Note that the $\max\{\cdot,\cdot\}$ operation is performed in component-wise manner. \textcolor{red}In particular, the input of the first layer is given by
\begin{align}
    \bm{p}^{0} & = \bm{z} - \bm{R}
    \begin{bmatrix}
        \bm{0}_{(m-k) \times 1} \\
        \bm{x}^k
    \end{bmatrix}.
\end{align}
In this case, the input size and output size are both fixed, which are set to $n_0=m$ and $n_L=1$, respectively. Hence, the structure of FCNN can be summarized as
\begin{align}\label{Eq::DNNStructure}
    \left\{m, n_1, n_2, \cdots, n^l, \cdots, n^{L-1}, 1\right\},
\end{align}
and the set of trainable parameters is given by
\begin{align}
    \bm{\theta} = \left\{ \bm{W}^1, \bm{b}^1, \cdots, \bm{W}^l, \bm{b}^l, \cdots, \bm{W}^L, \bm{b}^L \right\}.
\end{align}
We shall emphasize that the DNN's structure is not limited to FCNN, and it should be chosen according to the problem scale and resources. In this paper, we consider FCNN since it is very simple and can be easily implemented by hardware to improve the efficiency. Besides, FCNN will be a good candidate if we consider that the signal detection problem in MIMO systems has typically much lower dimensions with comparison to computer vision tasks like image reconstruction. It should be noted that the major objective of this paper is to show the power of DNN based heuristic, and the presented DNN structure in this work is just for reference. In practice, we may choose convolutional neural networks (CNNs) to further reduce the computational complexity when the problem scale is very large. On the other hand, recursive neural networks (RNNs) will be a good candidate if the channel is correlated over time. Therefore, we need to select an appropriate network structure to approximate the heuristic according to the specific applications and available resources for the considered system.

To leverage the DNN based heuristic function, we can combine it with heuristic BeFS search algorithms. Since A* may require prohibitively large memory space in large-scale problems, we will use the SMA* algorithm  \cite{Russell92} as the underlying search strategy. As a variant of A*, SMA* is able to perform BeFS with limited memory space, and it is still equivalent to A* with enough memory size. By combining SMA* with the aforementioned deep heuristic function, we thus propose a hyper-accelerated tree search (HATS) algorithm for the optimal signal detection in large-scale MIMO problems. The pseudo code of HATS is detailed in Algorithm \ref{Algo::HATS}, and the associated utility functions are presented in Algorithm \ref{Algo::HATSHelper}. In general, HATS works just like the A* algorithm. It keeps expanding the deepest least-$f$-cost node until \textsf{ACTIVE} is full, and the $f$-cost of each generated successor is computed according to (\ref{Eq::DeepCost}). In particular, since the memory size is limited, HATS will forget the most unpromising node from \textsf{ACTIVE}, and remember its key information in its parent. Thus, the memory size can be bounded. After safely deleting the most unpromising node, the proposed algorithm is able to move forward, and recover back once there is no any other path better than the forgotten paths. Based on this strategy, HATS maintains a partially expanded sub-tree of the whole tree. Hence, the most unpromising node shall be the shallowest highest $f$-cost leaf node. It should be noted that a node is said to be a ``leaf'' node in terms of the partially expanded sub-tree. Thus, the forgotten leaf node is not necessary to be the goal node. While maintaining the partially expanded sub-tree, the proposed algorithm will recursively adjust the expanded node's cost according to the costs of its successors, and thereby the sub-tree is updated. In conclusion, at each iteration, the proposed algorithm expands the best node from memory, generates one successor and inserts the generated successor to \textsf{ACTIVE} at the time, and deletes the worst leaf node when the memory is full.

\begin{algorithm}[t!]
    \caption{Proposed HATS Algorithm}\label{Algo::HATS}
    \hspace*{0.02in} { \bf Input:}{$\bm{z}$ and $\bm{R}$} \\
    \hspace*{0.02in} { \bf Output:}{$\hat{\bm{x}}$ (estimate of the transmitted signal $\bm{x}$)}
    \begin{algorithmic}[1]
        \State Insert the root node $\bm{x}^0$ into \textsf{ACTIVE}
        \Loop
            \If{\textsf{ACTIVE} is empty}
                \State \textbf{return} $\varnothing$ with failure
            \EndIf
            \State $\hat{\bm{x}}^k \gets $ \emph{deepest least-cost node} in \textsf{ACTIVE}
            \If{$\bm{x}^k$ is a goal node ($k=N_t$)}
                \State \textbf{return} $\hat{\bm{x}}^k$ with success
            \EndIf
            \State $\hat{\bm{x}}^{k+1} \gets$ next not-generated valid successor or best forgotten successor of {$\hat{\bm{x}}^k$}
            \State Insert $\hat{\bm{x}}^{k+1}$ into its parent's generated successor list
            \If{$\hat{\bm{x}}^{k+1}$ is not a forgotten node}
                \State $f(\hat{\bm{x}}^{k+1}) \gets \max\left(f(\hat{\bm{x}}^k), f_\theta(\hat{\bm{x}}^{k+1}\right))$
            \Else
                \State Recover $\hat{\bm{s}}^{k+1}$'s cost from its parent
            \EndIf
            \State \Call{Adjust}{$\hat{\bm{x}}^k$}
            \State \Call{MakeSpace}{$ $}
            \State Insert $\hat{\bm{x}}^{k+1}$ in \textsf{ACTIVE}
            \If{$\hat{\bm{x}}^k$ is expanded}
                \State remove $\hat{\bm{x}}^k$ from \textsf{ACTIVE}
            \EndIf
        \EndLoop
    \end{algorithmic}
\end{algorithm}

\begin{algorithm}[t!]
    \caption{Utility Functions for HATS}\label{Algo::HATSHelper} \label{Alg::SearchUtility}
    \begin{algorithmic}[1]
        \Function{Adjust}{$\hat{\bm{x}}^k$}
        \If{all of $\hat{\bm{x}}^k$'s successors are generated}
        \State $\hat{\bm{x}}^{k+1} \gets$ least-cost successor among all generated successors and forgotten successors of $\hat{\bm{x}}^{k+1}$
        \If{$\hat{\bm{x}}^{k+1}$'s cost is finite and not equal to $\hat{\bm{x}}^k$'s cost}
        \State Update $\hat{\bm{x}}^k$'s cost to $\hat{\bm{x}}^{k+1}$'s cost
        \State \Call{Adjust}{$\hat{\bm{x}}^{k}$'s parent}
        \EndIf
        \EndIf
        \EndFunction
        \Statex
        \Function{MakeSpace}{$ $}
        \If{\textsf{ACTIVE} is not full}
        \State \textbf{return}
        \EndIf
        \State Remove \emph{shallowest highest-cost leaf node} $\hat{\bm{x}}^k_j$ from \textsf{ACTIVE}
        \State Remove $\hat{\bm{x}}^k_j$ from its parent's generated successor list
        \State Remember $\hat{\bm{x}}^k_j$'s cost in its parent's forgotten successor list
        \If{the parent is not in \textsf{ACTIVE}}
        \State Insert the parent into \textsf{ACTIVE}
        \Call{MakeSpace}{}
        \EndIf
        \EndFunction
    \end{algorithmic}
\end{algorithm}

\subsection{Training Strategy}
It is clear that the training strategy is to minimize the average error between the estimated heuristic and the optimal one. To achieve this goal, we first know that with the optimal heuristic, the $f$-costs of the nodes on the shortest path of each sub-tree are all equal to the cumulative cost of the deepest node on that path. This can be easily proven according to the definition of the optimal heuristic. Following this point, one straightforward training strategy is to minimize the average $\ell_2$ loss
\begin{align}\label{Eq::LossMSE}
    \mathcal{L}(\bm{\theta}|\mathcal{D}) = \mathbb{E}_{\mathcal{D}} \left\{ \left\lvert g(\bm{x}^m_\phi) - f_{\bm{\theta}}(\bm{x}^k_\phi)\right\rvert^2 \right\},
\end{align}
where the empirical expectation is taken on the data set $\mathcal{D}$, and $\mathcal{D}$ can be generated by traversing the shortest path at each time slot. Specifically, the data set of $T$ time slots can be expressed as
\begin{align}
    \mathcal{D} = \left\{S^1_1, \cdots, S^{m-1}_1, \cdots, S^1_t, \cdots, S^k_t, \cdots, S^{m-1}_T\right\},
\end{align}
where $S^k_t$ denotes the $k$-th ($1 \leq k \leq m-1$) sample drawn at time slot $t$ ($1 \leq t \leq T$). Specifically, $S^k_t$ includes
\begin{align}
    \mathcal{S}^k_t = \left\{\bm{z}, \bm{R}, \bm{x}^k_\phi, \bm{x}^m_\phi\right\},
\end{align}
where $\bm{z}$, $\bm{R}$, $\bm{x}^m_\phi$ are the associated received signal, triangular matrix and ML estimate at the time. Notation $\bm{x}^{k}_\phi$ is the $k$-th node on the shortest path $\bm{x}^0 \rightarrow \bm{x}^m_\phi$. It should be noted that the time index is only used for identifying the symbols randomly drawn from different trees. After successfully collecting enough samples from different time slots, we will further shuffle the whole data set before training the model. Hence, the time index has no impact on the final prediction. Base on the data set, we can optimize the network parameters $\bm{\theta}$ by employing mini-batch gradient descent methods (e.g. the SGD optimizer  \cite{WerfelXS03} and ADAM optimizer  \cite{KingmaB14}) to find the optimal parameters as
\begin{align}
    \bm{\theta}^* = \mathop{\arg\min}_{\bm{\theta}} \mathcal{L}(\bm{\theta}|\mathcal{D}).
\end{align}

Nevertheless, we still have to compute the ML estimate, which will slow down the sampling efficiency during the training. To tackle this problem, we can use the transmitted vector $\bm{x}$ rather than the ML estimate $\bm{x}^m_\phi$. This is a reasonable approximation in the sense of signal estimation, since the ML estimate is most likely to be $\bm{x}$, especially for high SNR regimes. By using this strategy, we can train the model well very soon, and still have a very good performance. In summary, the advanced training procedure for the proposed algorithm is detailed in Algorithm \ref{Algo::Train}.
\begin{algorithm}[t!]
\begin{algorithmic}[1]
    \State // Generate the training set
    \State Set the number of mini-batches as $B$
    \State Set the number of time slots for each mini-batch as $T$
    \State initialize the data set $\mathcal{D}_{all}=\emptyset$
    \For{$j = 1, 2, \cdots, B$}
        \For{$t = 1, 2, \cdots, T$}
            \State Randomly collect $\bm{z}$, $\bm{R}$ and $\bm{x}$ with random SNR
            \State // Traverse the approximately shortest path of the whole tree
            \For{$k = 1, 2, \cdots, m$}
                \State Select the $k$-th node $\bm{x}^k$ on the path $\bm{x}$
                \State Collect $S^k_t = \{\bm{z}, \bm{R}, \bm{x}, \bm{x}^k\}$
                \State $\mathcal{D}_{all}=\mathcal{D}_{all} \cup \{S^k_t\}$
            \EndFor
        \EndFor
    \EndFor
    \State // Training with random samples
    \State Randomly initialize the DNN's parameters $\bm{\theta}$
    \For{$j=1, 2, \cdots, B$}
        \State Randomly select a mini-batch of samples $\mathcal{D}$ from $\mathcal{D}_{all}$
        \State Compute the average loss on $\mathcal{D}$ according to (\ref*{Eq::LossMSE})
        \State Perform a stochastic gradient decent step on the network's parameters $\bm{\theta}$
    \EndFor
    \State Save the well-trained parameters $\bm{\theta}$
\end{algorithmic}
\caption{Training Procedure for HATS}\label{Algo::Train}
\end{algorithm}

\subsection{Computational Complexity}
The computational complexity of HATS depends on both the number of visited nodes and the visitation costs. While visiting a node $\bm{x}^k$, the computational complexity is
\begin{align}
    \mathcal{O}\left(k + \sum_{l=1}^L n_l n_{l-1} + n_l\right),
\end{align}
where the first term denotes the complexity of computing $g(\bm{x}^k)$ and the second term denotes the complexity of computing $h_{\bm{\theta}}(\bm{x}^k)$. When the DNN estimates the optimal heuristic precisely, the algorithm visits the fewest nodes. Thus, the lower bound on the average complexity of HATS is given by
\begin{align}
    \mathcal{O}\left(m^2 + m\left(\sum_{l=1}^L n_l n_{l-1} + n_l\right)\right),
\end{align}
since only the $m$ nodes lying on the shortest path would be eventually expanded. When the estimation is imperfect, the optimality and the optimal efficiency can not be guaranteed, and the complexity may grow exponentially with the problem scale in the worst case. However, as we have discussed in Sec. \ref{Sec::OptimalEfficiency}, decreasing the estimation error will significantly improve the BER performance and search speed. Therefore, the performance and complexity of HATS rely on both the estimation quality of DNN and the system's SNR. In fact, as we will show in the simulations, the proposed algorithm achieves almost the optimal BER performance, while it reaches almost the lowest complexity in low SNR regimes. That is, with the proposed training strategy, the network is able to predict the optimal heuristic accurately. Thus, HATS can often reach the optimal efficiency under practical scenarios, which indicates that the proposed algorithm is applicable in large-scale systems.

\section{Simulation Results}
In this section, we will present simulation results and discussions to show the effectiveness of the proposed algorithm. Specifically, we will first introduce the environment setup and the implementation details of our model. Then, we will introduce the competing algorithms and discuss the related simulation results afterwards.

\subsection{Environment Setup}
In the simulations, we consider a MIMO system model where there are $m_c$ and $n_c$ antennas at the transmitter and receiver, respectively. Therefore, we have $m=2m_c$ and $n=2n_c$. In addition, the signal is modulated by QPSK or $16$-QAM, and the transmission experiences a random wireless channel. Moreover, the channel information can be known perfectly at the receiver. In practice, the columns of the CSI matrix may be somehow correlated. To perform simulations over correlated channels, we adopt the well-known Kroneker model introduced in  \cite{oestges2006validity} to generate the correlated channel matrices. Mathematically, the complex channel matrix $\bm{H}_c$ in Kroneker model is given by
\begin{align}
    \bm{H}_c = \bm{R}_r^{\frac{1}{2}}\tilde{\bm{H}}_c\bm{R}_t^{\frac{1}{2}},
\end{align}
where $\tilde{\bm{H}}_c$ denotes a Rayleigh flat fading channel matrix with independent and identically distributed (i.i.d.) random entries, and $\bm{R}_r$ and $\bm{R}_t$ are two covariance matrices at the receiver and transmitter, respectively. Without lost of generality, we assume that the correlation occurs at the transmitter, and $\bm{R}_r$ and $\bm{R}_t$ are set as
\begin{align}
    \bm{R}_r = \bm{I}, \quad \bm{R}_t =
    \begin{pmatrix}
    1       & \rho      & \cdots    & \rho      \\
    \rho    &   1       & \cdots    & \rho      \\
    \vdots  & \vdots    & \ddots    & \vdots    \\
    \rho    & \cdots    & \rho      & 1
    \end{pmatrix},
\end{align}
where $\rho \in [0, 1]$ is the correlation coefficient.

As to the DNN structure, we use the same DNN structure for all experiments, where there are $4$ hidden layers in total and the neuron numbers of the four hidden layers are set to $128$, $64$, $32$, and $16$, respectively. According to (\ref{Eq::DNNStructure}), the structure of DNN can be summarized as
\begin{align}
    \left\{ 2N_t, 128, 64, 32, 16, 1\right\}.
\end{align}
To train the model, we employ the Adam optimizer  \cite{KingmaB14} with the learning rate of $10^{-6}$ for all experiments. The mini-batch size is set to $128$ time slots, and the total batch size is $10$ million in the training. The training samples are randomly generated within a range of SNRs varying from $0$ dB to $30$ dB. For the performance test, we will test the aforementioned algorithms with enough times to ensure a stable BER performance.

\subsection{Competing Algorithms}
To verify the effectiveness of the proposed algorithm, we  compare the proposed algorithm with several competitive algorithms. Before discussing the results, we first introduce the following abbreviations,
\begin{itemize}
    \item \emph{MMSE}: The conventional minimum mean squared error estimator.
    \item \emph{OAMP-Net2}: The orthogonal approximate message passing network 2 introduced in  \cite{HeWJL20}.
    \item \emph{SD}: The sphere decoding algorithm introduced in  \cite{HassibiV05}.
    \item \emph{DL-SD}: The DL based sphere decoding algorithm introduced in  \cite{MohammadkarimiM19}.
    \item \emph{A*}: The A* algorithm introduced in  \cite{Russell92}. Note that the heuristic is set to zero constantly for comparison.
    \item \emph{HATS($M$)}: The proposed algorithm in this paper, where $M$ denotes the memory capacity.
\end{itemize}
Among these algorithms, OAMP-Net2, DL-SD and the proposed HATS algorithm are DL based algorithms. Besides, SD, DL-SD, A* and HATS are search algorithms. In particular, DL-SD requires linear space with the search depth, while A* has exponential space complexity. Moreover, we will take simulations for HATS with different memory sizes. In particular, HATS($\infty$) denotes that the memory size is unlimited. In this case, HATS($\infty$) is equivalent to A* except that the heuristic is estimated by a DNN. Since SD, DL-SD and A* are optimal search algorithms, we will simply denote their BER performances as ``ML'' for convenience. Unlike the existing literature that measures the complexity in terms of the number of visited lattice points (goal nodes)  \cite{MohammadkarimiM19}, we will compare the complexities of search algorithms based on the number of visited nodes in the simulations, since it is more accurate to measure the complexity.

\subsection{BER Performance and Complexity Comparison On Uncorrelated Channels}
\begin{figure}[t!]
    \centering
    \includegraphics[width=3.25in]{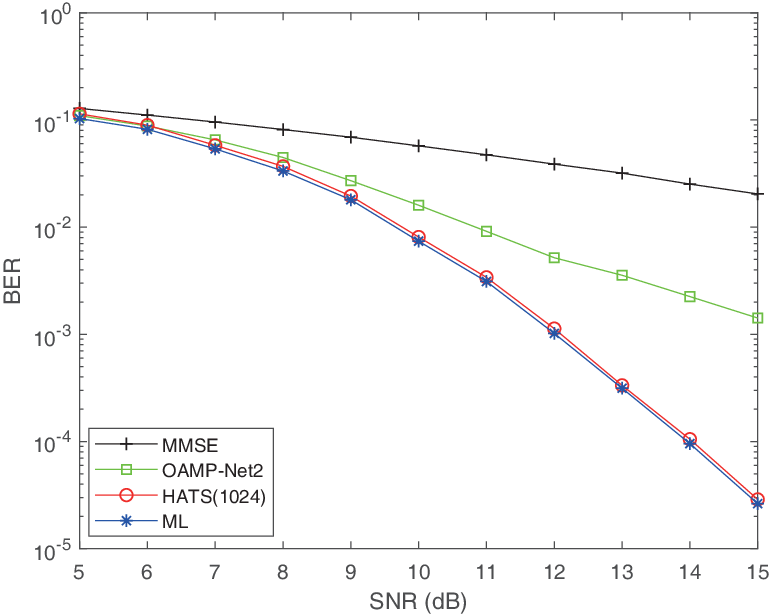}
    \caption{BER comparison versus SNR for $8 \times 8$ MIMO with uncorrelated channels.}
    \label{Fig::BER_vs_snr_QPSK8}
\end{figure}

\begin{figure}[t!]
    \centering
    \includegraphics[width=3.25in]{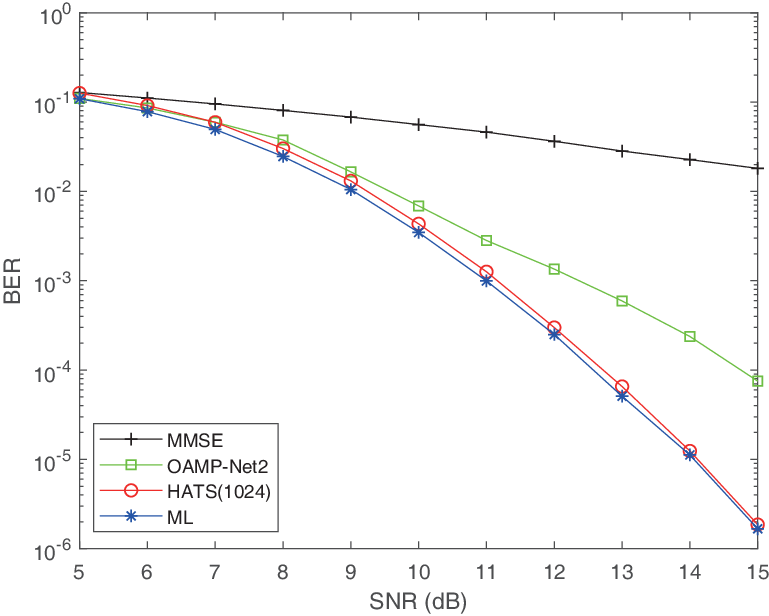}
    \caption{BER comparison versus SNR for $16 \times 16$ MIMO with uncorrelated channels.}
    \label{Fig::BER_vs_snr_QPSK16}
\end{figure}

\begin{figure}[t!]
    \centering
    \includegraphics[width=3.25in]{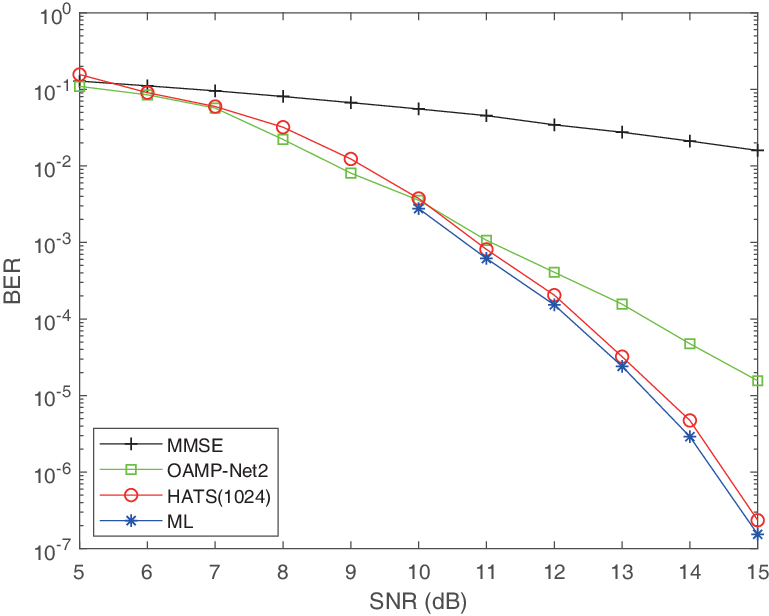}
    \caption{BER comparison versus SNR for $32 \times 32$ MIMO with uncorrelated channels.}
    \label{Fig::BER_vs_snr_QPSK32}
\end{figure}

Figs. \ref{Fig::BER_vs_snr_QPSK8}-\ref{Fig::BER_vs_snr_QPSK32} illustrate the BER comparisons of the aforementioned algorithms in different-scale MIMO systems, where the channel is uncorrelated, and the numbers of antennas are $8$, $16$, $32$ at both the transmitter and receiver, respectively. In addition, SNR varies from $5$ dB to $15$ dB. Note that algorithms have to search the closest lattice point among $2^{64}$ candidates to reach the exact optimal BER performance in $32 \times 32$ MIMO, which is a very big challenge. It should be noted that all the ML performances in these figures are simulated based on the SD algorithm. In particular, the BER results of ``ML'' with SNR ranging from $5$ dB to $9$ dB are not plotted in Fig. \ref{Fig::BER_vs_snr_QPSK32}, since the computational complexity of SD becomes prohibitively large in this regime. From these figures, we can find that HATS achieves almost the optimal BER performance in all three systems. When $\text{SNR}=15$ dB, HATS produces only slightly $10\%$, $10\%$ and $15\%$ more error than the optimal search algorithms in $8 \times 8$, $16 \times 16$ and $32 \times 32$ MIMO systems, respectively. As a contrast, for the sub-optimal algorithms like MMSE and OAMP-Net2, their BER performances are significantly far from the optimal one in all three MIMO systems, especially when the problem scale becomes prohibitively large. Moreover, we can also find from these figures that the performance gaps between HATS and the other sub-optimal algorithms enlarge with the increasing SNR. In particular, when $\text{SNR}=15$ dB, HATS reduces the error of MMSE and OAMP-Net2 to only about $0.0052\%$ and $1.504\%$ in $32 \times 32$ MIMO, respectively. Besides, the SNR gains of HATS over OAMP-Net2 are about $3.2$ dB, $2.1$ dB and $1.8$ dB at the BER levels of $10^{-3}$, $10^{-4}$, and $10^{-5}$ in $8 \times 8$, $16 \times 16$ and $32 \times 32$ MIMO systems, respectively. In conclusion, these results show that the proposed algorithm is robust and almost optimal even in large-scale problems.

\begin{figure}[t!]
    \centering
    \includegraphics[width=3.25in]{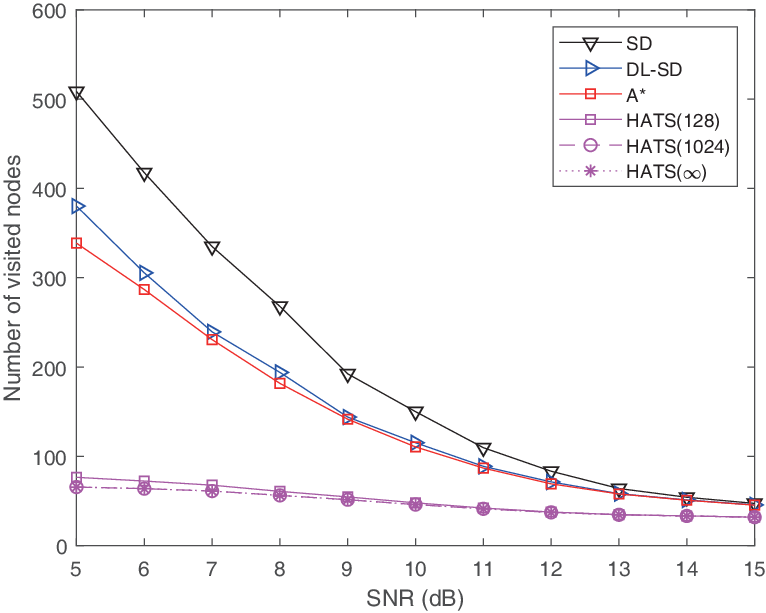}
    \caption{Complexity comparison versus SNR for $8 \times 8$ MIMO with uncorrelated channels.}
    \label{Fig::Cm_vs_snr_QPSK8}
\end{figure}

\begin{figure}[t!]
    \centering
    \includegraphics[width=3.25in]{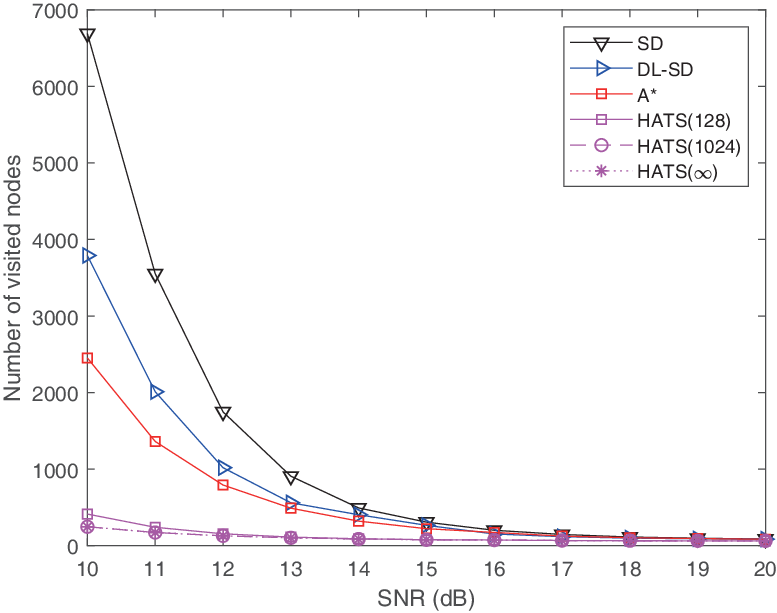}
    \caption{Complexity comparison versus SNR for $16 \times 16$ MIMO with uncorrelated channels.}
    \label{Fig::Cm_vs_snr_QPSK16}
\end{figure}

\begin{figure}[t!]
    \centering
    \includegraphics[width=3.25in]{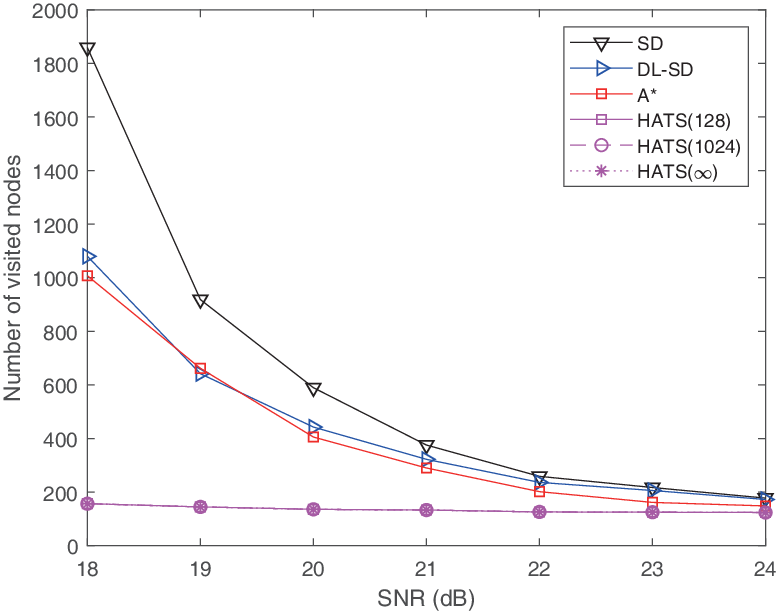}
    \caption{Complexity comparison versus SNR for $32 \times 32$ MIMO with uncorrelated channels.}
    \label{Fig::Cm_vs_snr_QPSK32}
\end{figure}

In order to show the impact of SNR and memory constraint on the computational complexity of the proposed algorithm, Figs. \ref{Fig::Cm_vs_snr_QPSK8}-\ref{Fig::Cm_vs_snr_QPSK32} are provided to show the complexity comparison under the same setup as Figs. \ref{Fig::BER_vs_snr_QPSK8}-\ref{Fig::BER_vs_snr_QPSK32}. From the results in Figs. \ref{Fig::Cm_vs_snr_QPSK8}-\ref{Fig::Cm_vs_snr_QPSK32}, we can conclude that with difference memory sizes, HATS still visits much fewer nodes than SD, DL-SD and A*. The numbers of nodes visited by SD, DL-SD and A* all grow exponentially with the decreasing SNR, which indicates that they are not applicable in large-scale problems. As a contrast, the number of nodes visited by HATS increases almost linearly in low SNR regimes. Moreover, we can find from these figures that HATS can reach the lowest complexity in a wide range of acceptable SNRs, whereas the complexities of SD, DL-SD and A* are still very high in the regime. Specifically, for the $32 \times 32$ MIMO where the size of search space is $2^{64}$ and $\text{SNR}=18$ dB, SD, DL-SD and A* have to visit average about $1800$, $1100$ and $1000$ nodes to find the optimal solution, while HATS only needs to visit about average $150$ nodes. In particular, we can also find that HATS is not sensitive to the memory size, and it can perform well with bounded memory size. Specifically, when the memory size is bounded, the performances of HATS($128$), HATS($1024$) and HATS($\infty$) are almost the same in all the three systems. This indicates that HATS can reach almost the optimal efficiency with bounded memory in practice.

\begin{figure}[t!]
    \centering
    \includegraphics[width=3.25in]{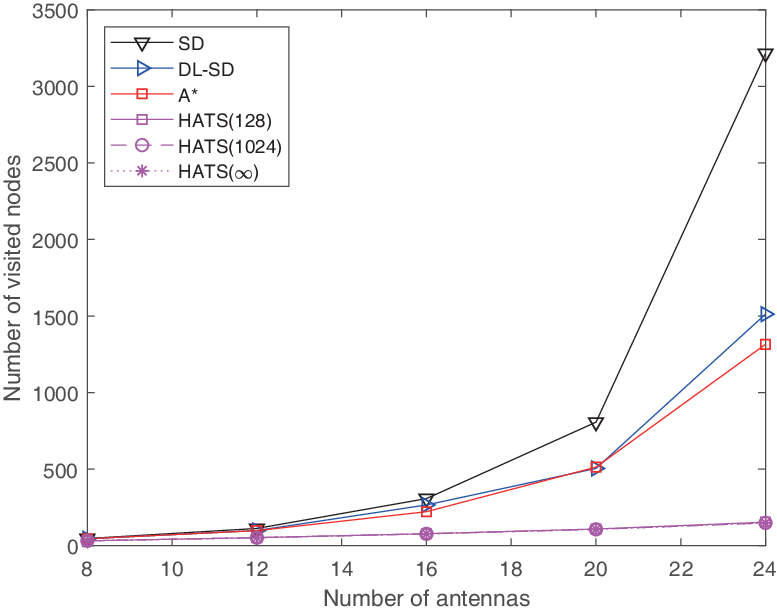}
    \caption{Complexity comparison versus number of antennas with uncorrelated channels ($\text{SNR}=15$ dB).}
    \label{Fig::Cm_vs_m}
\end{figure}

Fig. \ref{Fig::Cm_vs_m} illustrates the computational complexity comparison of the competing algorithms versus the problem scale, where the number of antennas varies from $8$ to $24$ and the associated SNR is set to $15$ dB. From this figure, we can observe that the complexities of HATS($128$), HATS($1024$) and HATS($\infty$) increase almost linearly with the number of antennas, while the complexities of SD, DL-SD, A* increase exponentially with the problem scale. This implies that HATS is much more efficient and less sensitive to the problem scale by comparing to the other optimal search algorithms. As it has been shown in Figs. \ref{Fig::BER_vs_snr_QPSK8}-\ref{Fig::Cm_vs_snr_QPSK32} that HATS can reach almost the optimal BER performance in large-scale systems, one is able to conclude that the proposed algorithm almost meets the optimal efficiency in large-scale problems, which indeed further verifies the effectiveness of the proposed algorithm.

\subsection{Further Results on Correlated Channels and Higher-Order Modulation}
\begin{figure}[t!]
    \centering
    \includegraphics[width=3.25in]{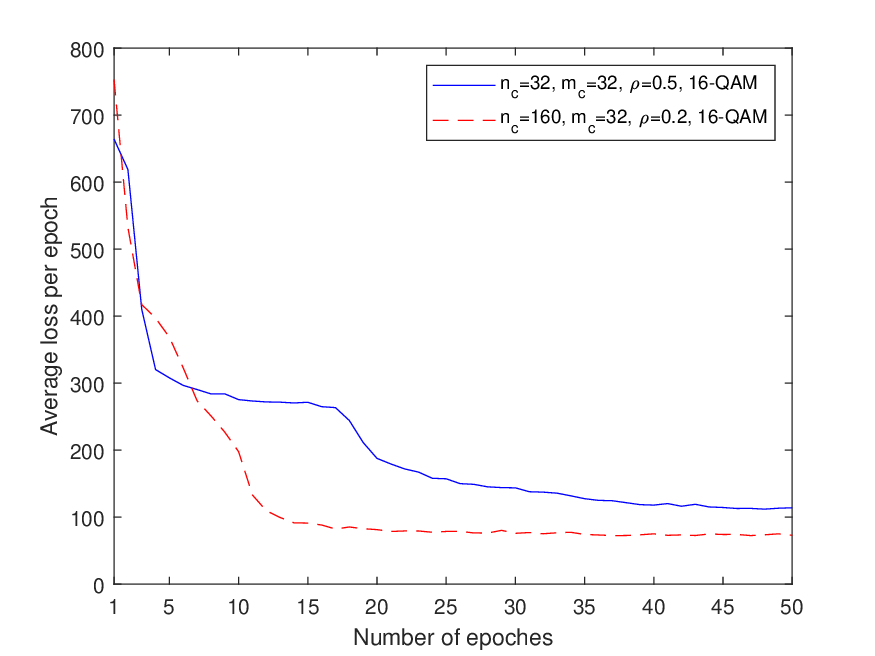}
    \caption{Training loss for $16$-QAM modulated MIMO systems with correlated channels.}
    \label{Fig::train_16qam}
\end{figure}

\begin{figure}[t!]
    \centering
    \includegraphics[width=3.25in]{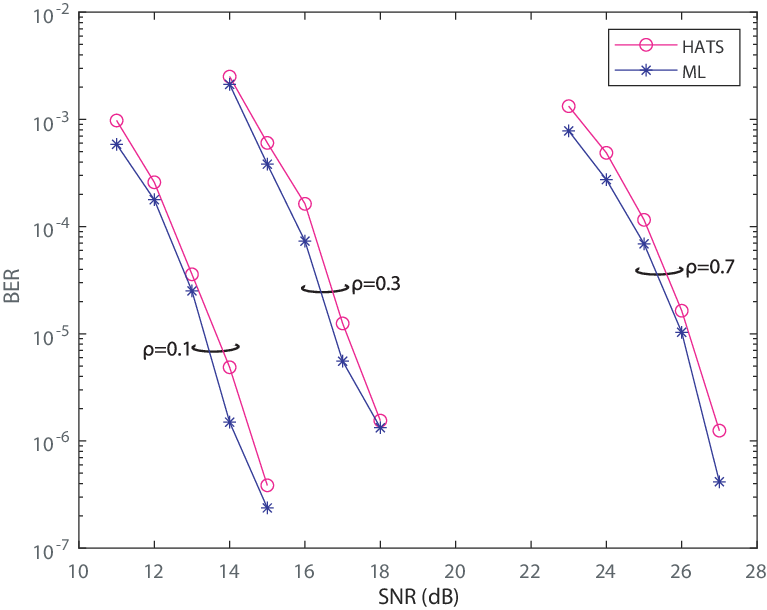}
    \caption{BER comparison versus SNR for $16$-QAM modulated MIMO systems with $m_c=32$, $n_c=160$, and correlated channels.}
    \label{Fig::ber_vs_snr_32x160_16qam}
\end{figure}

\begin{figure}[t!]
    \centering
    \includegraphics[width=3.25in]{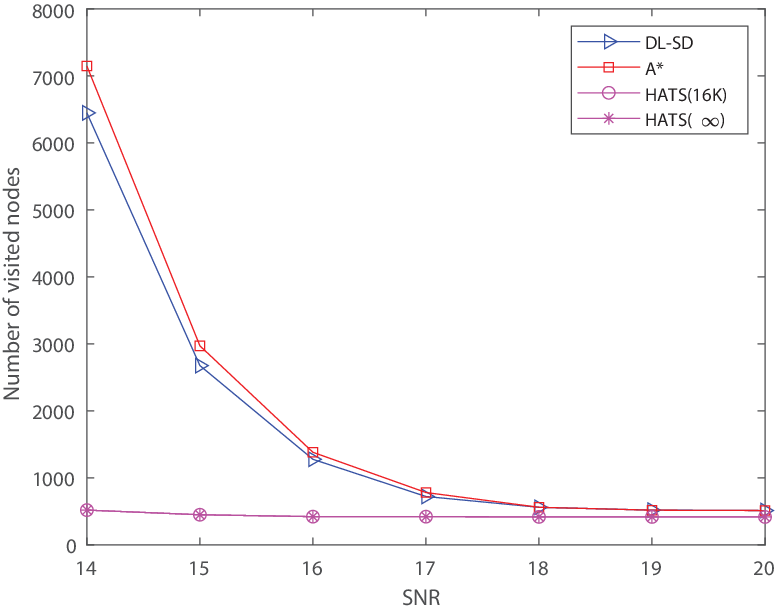}
    \caption{Complexity comparison versus SNR for $16$-QAM modulated MIMO systems with $m_c=32$, $n_c=160$ and $\rho=0.1$.}
    \label{Fig::complexity_vs_snr_32x160_16qam}
\end{figure}

\begin{figure}[t!]
    \centering
    \includegraphics[width=3.25in]{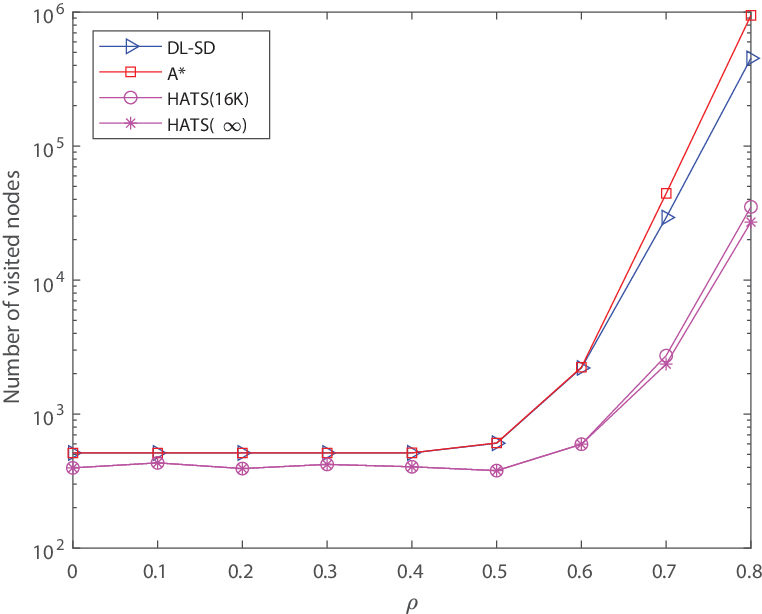}
    \caption{Complexity comparison versus $\rho$ for $16$-QAM modulated MIMO systems with $m_c=32$, $n_c=160$, SNR=$20$ dB.}
    \label{Fig::complexity_vs_rho_32x160_16qam}
\end{figure}

To show the robustness of the proposed HATS in practical scenarios, we further present simulation results in Figs. \ref{Fig::train_16qam}-\ref{Fig::complexity_vs_rho_32x160_16qam}, in which the channels are correlated ($\rho > 0$) and $16$-QAM is adopted. Specifically, Fig. \ref{Fig::train_16qam} demonstrates the average training loss of the proposed HATS on two different MIMO systems, where the numbers of transmit antennas are both $32$, while the numbers of receive antennas are $32$ and $160$, and the channel correlation coefficients are $0.5$ and $0.2$, respectively. We can find from Fig. \ref{Fig::train_16qam} that the average training losses of the proposed HATS rapidly converge to a low value in both of the two systems, which indicates that the proposed network structure and training strategy are effective for different practical systems.

Besides, Fig. \ref{Fig::ber_vs_snr_32x160_16qam} depicts the BER performance comparison results of the aforementioned algorithms, where the correlation coefficient $\rho$ varies in $\{0.1, 0.3, 0.7\}$, and SNR varies from $10$ dB to $30$ dB. From this figure, we can see that even when the size of search space grows very largely to $2^{128}$ for a $16$-QAM modulated $32 \times 160$ large-scale MIMO system, and the channels are slightly ($\rho=0.1$), moderately ($\rho=0.3$), or strongly ($\rho=0.7$) correlated, HATS can still be able to reach almost the optimal ML performance in a wide range of SNRs. This indicates that the estimated heuristic is nearly admissible such that the algorithm can always successfully find the ML estimate. Moreover, Fig. \ref{Fig::complexity_vs_snr_32x160_16qam} demonstrates the complexity comparison over SNR for the corresponding BER results of $\rho=0.1$ in Fig. \ref{Fig::ber_vs_snr_32x160_16qam}. We can find from this figure that the proposed HATS still outperforms the A* and DL-SD algorithms in a wide range of SNRs, and it visits nearly only the $512$ necessary nodes connecting the shortest path for this system. This indicates that the estimated heuristic is nearly optimal, by the fact that the proposed HATS not only achieves almost the optimal BER performance, but also reaches almost the optimal efficiency. In further, we present the complexity comparison results over the correlation level in Fig. \ref{Fig::complexity_vs_rho_32x160_16qam}, where the correlation coefficient $\rho$ ranges from $0$ to $0.8$. It can be seen from Fig. \ref{Fig::complexity_vs_rho_32x160_16qam} that the proposed HATS is not sensitive to the correlation level with comparison to the original A* and DL-SD algorithms. The proposed HATS will maintain the average complexity at a low level until the correlation level grows really high to $0.8$, while the other two algorithms' complexities will grow rapidly with the increasing correlation level. This indicates that the proposed model can still estimate a high-quality heuristic even when the channel is strongly correlated, which indeed further verifies the robustness of the proposed algorithm.

\section{Conclusions and Future Works}
In this paper, we investigated the optimal signal detection problem with focus on large-scale MIMO systems. The problem can be regarded as search on the decision tree, and the optimal solution is obtained by finding the shortest path. In order to improve the search speed while does not compromise the optimality, we proposed a DL based heuristic search algorithm, namely, HATS. The proposed algorithm was shown to be almost optimally efficient, since it can reach almost the optimal BER performance while still visit almost the fewest nodes in large-scale systems. Hence, we do believe that the proposed algorithm is attractive for optimal signal detection in practical large-scale MIMO systems, especially when the problem scales largely and the optimal performance is required.

Generally, this paper have focused mainly on producing hard-decision output. However, channel coding is often used in modern MIMO systems to enhance the reliability, and thereby having probabilistic soft-output will help enhance the performance of wireless transmission. Therefore, one interesting future topic of this work is to provide soft-decisions with HATS, and this could be a challenge for all tree search based algorithms. A feasible solution to address this challenge is to employ the max-log approximation to compute log-likelihood ratios (LLRs) of bits, such that we will only need to visit the nodes neighbouring to the shortest path  \cite{PrasadKW11}. In this way, we can easily compute LLRs of bits by designing an additional pruning strategy to neglect all the nodes which have no contributions to LLRs, which indicates that it becomes possible to provide soft-output while still visit as fewer nodes as possible.

\bibliographystyle{IEEEtran}
\bibliography{IEEEabrv,CRN}

\end{document}